\documentclass[11pt]{article}
 \pdfoutput=1

\usepackage{arxiv}

\usepackage[utf8]{inputenc} 
\usepackage[T1]{fontenc}    
\usepackage{hyperref}       
\usepackage{url}            
\usepackage{booktabs}       
\usepackage{amsfonts}       
\usepackage{nicefrac}       
\usepackage{microtype}      
\usepackage{lipsum}		
\usepackage{graphicx}
\usepackage{natbib}
\usepackage{doi}
\usepackage{xcolor}
\usepackage{tcolorbox}
\usepackage{subfiles}
\usepackage{geometry}
\usepackage{array}
\usepackage{color}
\usepackage{soul}
\usepackage{graphicx}
\usepackage{float}
\graphicspath{ {images/} }

\title{The social impact of Generative AI: An Analysis on ChatGPT}

\date{September, 2023}	

\author{ \href{https://orcid.org/0000-0001-8589-2850}{\includegraphics[scale=0.06]{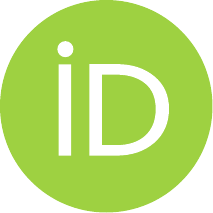}\hspace{1mm}Maria T. Baldassarre} \\
	Department of Computer Science\\
	University of Bari "A. Moro"\\
	Bari, BA 70125\\
	\texttt{mariateresa.baldassarre@uniba.it} \\
	\And
	\href{https://orcid.org/0000-0001-5719-7447}{\includegraphics[scale=0.06]{orcid.pdf}\hspace{1mm} Danilo Caivano} \\
	Department of Computer Science\\
	University of Bari "A. Moro"\\
	Bari, BA 70125 \\
	\texttt{danilo.caivano@uniba.it} \\
    \And
	\href{https://orcid.org/0009-0007-0468-5050}{\includegraphics[scale=0.06]{orcid.pdf}\hspace{1mm} Berenice Fernández Nieto} \\
	Department of Computer Science\\
	University of Bari "A. Moro"\\
	Bari, BA 70125 \\
	\texttt{berenice.fernandeznieto@uniba.it} \\
    \And
	\href{https://orcid.org/0000-0003-3589-6970}{\includegraphics[scale=0.06]{orcid.pdf}\hspace{1mm} Domenico Gigante} \\
	Department of Computer Science\\
	University of Bari "A. Moro"\\
	Bari, BA 70125 \\
	\texttt{domenico.gigante1@uniba.it}\\
    \And
	\href{https://orcid.org/0000-0002-3537-7663}{\includegraphics[scale=0.06]{orcid.pdf}\hspace{1mm} Azzurra Ragone} \\
	Department of Computer Science\\
	University of Bari "A. Moro"\\
	Bari, BA 70125 \\
	\texttt{azzurra.ragone@uniba.it}\\
}

\renewcommand{\shorttitle}{\textit{The Social Impact of Generative AI}}

\hypersetup{
pdftitle={A template for the arxiv style},
pdfsubject={q-bio.NC, q-bio.QM},
pdfauthor={David S.~Hippocampus, Elias D.~Striatum},
pdfkeywords={First keyword, Second keyword, More},
}

\begin{document}
\maketitle

\begin{abstract}
In recent months, the social impact of Artificial Intelligence (AI) has gained considerable public interest, driven by the emergence of Generative AI models, ChatGPT in particular. The rapid development of these models has sparked heated discussions regarding their benefits, limitations, and associated risks. Generative models hold immense promise across multiple domains, such as healthcare, finance, and education, to cite a few, presenting diverse practical applications. Nevertheless, concerns about potential adverse effects have elicited divergent perspectives, ranging from privacy risks to escalating social inequality. This paper adopts a methodology to delve into the societal implications of Generative AI tools, focusing primarily on the case of ChatGPT. It evaluates the potential impact on several social sectors and illustrates the findings of a comprehensive literature review of both positive and negative effects, emerging trends, and areas of opportunity of Generative AI models. This analysis aims to facilitate an in-depth discussion by providing insights that can inspire policy, regulation, and responsible development practices to foster a human-centered AI.
\end{abstract}


\section{Introduction}
In recent months, the social impact of Artificial Intelligence (AI) has been at the forefront of public debate due primarily to the introduction of new software systems and technologies, specifically ChatGPT. The rapid development of these technologies has, even more, sparked the debate regarding the advantages, limitations, and risks of Artificial Intelligence's expanding capabilities. From healthcare to cybersecurity, generative models offer a vast array of practical and prosperous future possibilities. However, concerns regarding potential adverse effects present an opposing viewpoint, with arguments spanning from privacy risks to deepening social inequalities.
Models like ChatGPT personalize the digital version of the Delphic oracle, where people expect to find answers to their current problems by automating tasks, seeking ChatGPT's opinion on various issues, and even requesting advice. Nonetheless, it is essential to question whether we are genuinely resolving uncertainties or uncovering new ones regarding the scope, boundaries, and prospects of generative models' societal impact.
Some other analysts have referred to an "AI arms race" in which companies worldwide strive to showcase the best technology, innovation prowess, and leadership in the AI market. On the other side of the debate, discussions refer to the rapid development of these models and the effectiveness of existing legal frameworks in safeguarding against unintended adverse outcomes.
Amongst all these debates, unquestionably, Generative AI is currently undergoing a period of accelerated evolution. This evolution inevitably brings about a social impact akin to the ones experienced through numerous other technological advancements that transformed our society in the past. To date, significant effects have been observed in service provision, education, and scientific analysis. However, more profound and concerning impacts also unfold in domains like democracy, inequality, security, and military technology.

Consequently, a comprehensive examination and analysis are required to understand the positive and negative social consequences, emerging trends, and areas of improvement of generative models. These studies are needed to address potential vulnerabilities and ensure the development of these technologies considers the diverse social contexts and realities in which they are deployed.

Building upon the preceding insights, this analysis adopts a comprehensive approach to explore the societal ramifications and future trajectories of Generative AI, with a specific emphasis on ChatGPT. The analysis is organized as follows: Section \ref{sec:background} examines the potential impacts of ChatGPT across diverse social sectors and the evolution of the debate across various spheres; Section \ref{sec:Steteof} provides a brief overview of the state-of-the-art of Generative AI models as well as a classification of these; Section \ref{sec:studydesign} presents the study design, goals, and Research Questions, plus the search strategy. Section \ref{sec:Dataanalysis} includes the data analysis and synthesis, drawing conclusions from the literature review, and sharing visualizations of our findings. Conclusion and future work close the paper.

\section{BACKGROUND}
\label{sec:background}
Throughout history, the advancement of technology has consistently brought about significant transformations in social dynamics. Each new technological breakthrough has sparked debates regarding scientific progress' advantages and potential hazards. Currently, this discourse encompasses various automated tools, data collection and analysis, and the digitization of services, among other emerging applications, which have become integral parts of modern-day life.

These novel technological applications pervade numerous societal domains, ranging from education to diplomacy, exerting a profound influence and continuously reshaping various social processes. While there is a prevailing and optimistic belief in the positive impact of technology on human progress \citep{mckendrick_learning_nodate}, it is also becoming increasingly apparent that these disruptive forces could potentially engender unforeseen and unintended consequences.

In informatics, sociology, philosophy, and politics, the development of generative models will continue to ignite in-depth discussions on various subjects. These discussions include topics such as regulation, risk mitigation, liability, transparency, and accountability, as well as the effects on socialization patterns and the trajectory of technological development itself.

In our case, the significance of examining the social impact of ChatGPT stems from its potential to cause significant social transformations, despite ongoing debates regarding the magnitude of these changes. ChatGPT is a powerful generative model that may impact power dynamics at multiple scales, from individual interactions to broader social structures \citep{farrell_spirals_2022}.
This dynamic occurs in complex social environments marked by disparities, stereotypes, conflicts, and various political and social organization forms. These diverse social contexts, which surround scientific advances, trigger unpredictable and immeasurable consequences that fundamentally transform how we interact with each other and the world. 
When a disruptive force permeates a new environment, it encounters various forms of resistance. It triggers unintended negative consequences, demands protection from potential vulnerabilities, causes uncertainty about whether it can be regulated, and other related factors. This critical-resistant front contrasts with a skeptical perception that questions the gravity of this new disruption and the alarmist interpretations surrounding emerging technologies, in this case. A third front embodies an optimistic outlook, emphasizing the manifold benefits across multiple sectors, fostering enthusiasm for potential advancements and enhancements, and envisioning their potential to address significant challenges. Overall, the interaction between these forms of resistance shapes perceptions of generative models' societal impact and raises critical questions about their implications for the broader social fabric.

These perspectives, as human behavior changes over time, interact and mutually influence and foster one another. In the case of ChatGPT, we also observe these attitudes, including optimism, pessimism, and skepticism, but the panorama is even more complex as we present below in our literature review. 

As evidence of the increasing interest in ChatGPT and AI, Figure 1 depicts the fluctuations in Google search queries for "ChatGPT" globally from May 2020 (the release of ChatGPT) until May 2023 (which includes the time frame of our research).
\begin{figure}[H]
    \centering
    \includegraphics[scale=0.40]{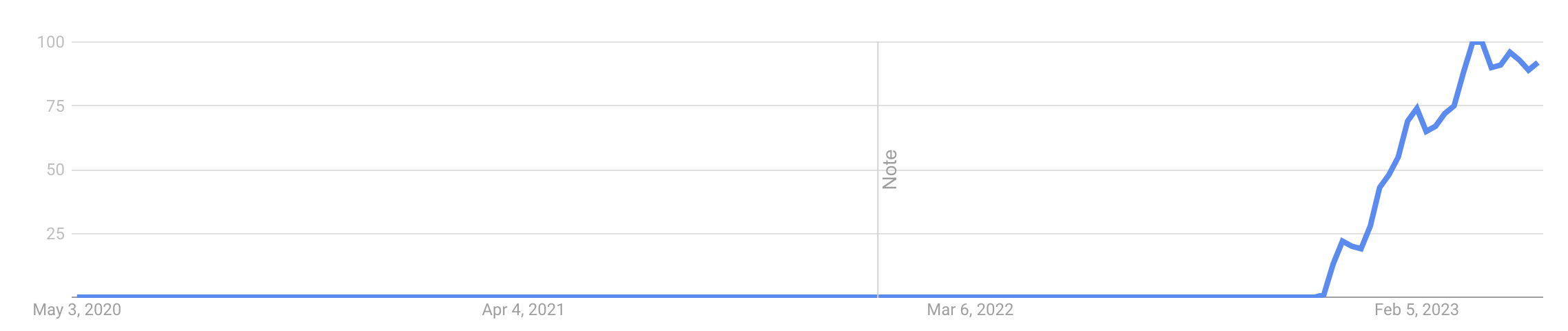}
    \caption{Search queries evolution on “ChatGPT” from May 2020 to May 2023 via Google Trends (\url{https://trends.google.it/trends/explore?date=2020-04-28\%202023-05-12\&q=Chat\%20GPT3\&hl=en}). The "note" in the graph reflects an improvement in Google's data collection system implemented on 1 January 22.}
    \label{fig:Google1}
\end{figure}

Fig.\ref{fig:Google1}, furthermore, depicts a consistent low level of interest until December 2022, when a significant shift occurred, coinciding with the months following the launch of Open AI's ChatGPT as a prototype service on 30 November 2022 \citep{gordon_will_2022}, which attracted global attention. In April and May of 2023, interest peaked.



As we enter the third wave of AI evolution, examining how socialization processes change and assessing scientific innovations' potential positive or negative effects on rights and freedoms is critical. Following Pinch and Bijker \citep{pinch_social_1984}, it is essential to examine the construction of scientific knowledge across different localities and contexts. As a result, the primary goal of this analysis is to assess the evolution of ChatGPT in recent years and to examine its current perceived impact on various social aspects, in the context of the ongoing wave of AI evolution.


\section{STATE OF THE ART}
\label{sec:Steteof}
\textbf{Generative pre-training} (GP) was a well-known concept in machine learning applications since 2012 \citep{noauthor_acoustic_2012}. Later, in 2017 Google introduced the transformer architecture \citep{vaswani_attention_2017}. 
These advancements led to the birth of large language models like \textit{BERT} in 2018 \citep{devlin_bert_2019} and \textit{XLNet} in 2019 \citep{yang_xlnet_2019}: these are pre-trained transformers (PT) but are not designed to be generative. 
A \textbf{language model} is a probability distribution over sequences of words \citep{jurafsky_speech_2008}: given any sequence of words of length \textit{m}, a language model assigns a probability to the whole sequence. A \textbf{Large Language Model} (LLM) is a language model based on a neural network with many parameters (typically billions or more).
Prior to transformer-based architectures, the best-performing neural Natural Language Processing (NLP) models employed supervised learning from large amounts of manually-labeled data.

The main drawbacks of using supervised learning are the impossibility to use it on not well-annotated datasets, and also the prohibitive cost and time required to train extremely large language models \citep{Radford2018ImprovingLU}. 
Usually, LLMs trained on a large quantity of data can perform discretely a good number of tasks; anyway, they can be fine-tuned (i.e., further trained on specific data) to execute a specific task with better performance.

Later, in 2018 OpenAI \citep{openai_ref} published its famous article "\textit{Improving Language Understanding by Generative Pre-Training}", in which the first \textbf{Generative Pre-trained Transformer} (GPT) system was introduced \citep{Radford2018ImprovingLU}. 
GPT is a type of large language model (LLM) used mainly for Generative AI, which is a type of AI capable of generating various kinds of content, such as text and images, in response to instructions (also known as prompts). Generative AI models learn the patterns and structure of their input training data and then generate new data that has similar characteristics, according to what has been asked as a prompt.\\

\subsection{Foundational models}

A foundational model is an AI model trained on broad data at scale such that it can be adapted to a wide range of downstream tasks. 
The most famous and performant GPT foundation models are the ones released by OpenAI. The most recent is \textbf{GPT-4} \citep{gpt4_ref}, for which OpenAI refused to publish the size and training details due to business reasons \citep{openai_refuse_ref}.


Other such models include Google's \textbf{PaLM} \citep{palm2_ref} and Together's \textbf{GPT-JT} \citep{gptjt_ref}, which has been reported as the closest-performing open-source alternative to GPT-3. Meta as well has released a generative foundational language model, named \textbf{LLaMA} \citep{llama_ref}.
Foundational GPTs can also handle media (and not only text), both for input and/or output. For example, GPT-4 is capable of processing text and images as input but only produces text as output. 

\subsection{Task-Specific Models}

A foundational GPT model is usually further trained to better perform specific tasks and/or handle subject-matter domains. One of the most used methods for such adaptation is fine-tuning (beyond that done for the foundation model). Fine-tuning is an approach in which the weights of a pre-trained (language) model are trained on new data.
One example of this is fine-tuning LLM to comprehend and follow instructions: in January 2022, OpenAI introduced \textbf{InstructGPT} \citep{instructgpt_ref} --a series of models which were fine-tuned to follow instructions. The gained advantages included higher accuracy, less negative/toxic sentiment, and generally better alignment with user needs. 
Other examples of task-specific models are chatbots, AI systems that engage in human-like conversation; \textbf{ChatGPT} \citep{chatgpt_ref} is currently the most famous chatbot. Anyway, other major chatbots currently exist, such as Microsoft's \textbf{Bing Chat} \citep{bing_ref} -- which uses OpenAI's \textbf{GPT-4} \citep{gpt4_ref} (as part of a broader close collaboration between OpenAI and Microsoft) --, Google's competing chatbot \textbf{Bard} \citep{bard_ref} and \textbf{LaMDA} \citep{lamda_ref}, \textbf{Jasper Chat} \citep{jasper_ref}, \textbf{Claude} \citep{claude_ref}.

Finally, the text-to-model task is becoming quite popular. To date, some famous models are \textbf{Dall-E 2} \citep{dalle2_ref}, \textbf{Stable Diffusion} \citep{stable_diffusion_ref}, \textbf{PhotoSonic Art Generator} \citep{photosonic_art_ref} whose task is the production of images based on user-provided textual prompts.
Following the text-to-image models, also the text-to-video task has been addressed with a lot of models, such as: \textbf{Runway} \citep{runway_ref}, Meta’s \textbf{Make-A-Video} \citep{makeavideo_ref}, Google’s \textbf{Imagen Video} \citep{imagen_ref} and \textbf{Phenaki} \citep{phenaki_ref}. All these models can generate video from text and/or text/image prompts. 

\subsection{Domain-specific models}

GPT systems can be re-trained to address particular fields or domains. Examples of such models (and apps) are \textbf{BloombergGPT} \citep{bloomberg_gpt_ref} for the financial domain, which should provide help with financial news and information, \textbf{SlackGPT} \citep{slack_gpt_ref} to support the Slack instant-messaging service by providing help and guidance with navigating and summarizing discussions (based on OpenAI's API), \textbf{BioGPT} \citep{bio_gpt_ref} for the biomedical domain, to provide help with biomedical literature text generation and mining, \textbf{CoPilot} \citep{copilot_ref} for the IT source code development domain, to provide auto-completion capabilities for developers.

Sometimes domain-specificity is realized via software components, specifically named plug-ins or add-ons. For example, Google Workspace has available add-ons such as \textit{GPT for Sheets and Docs} -- which is reported to aid the use of spreadsheet functionality in Google Sheets. 

\section{STUDY DESIGN}
\label{sec:studydesign}

\begin{figure}[H]
    \centering
    \includegraphics[scale=0.19]{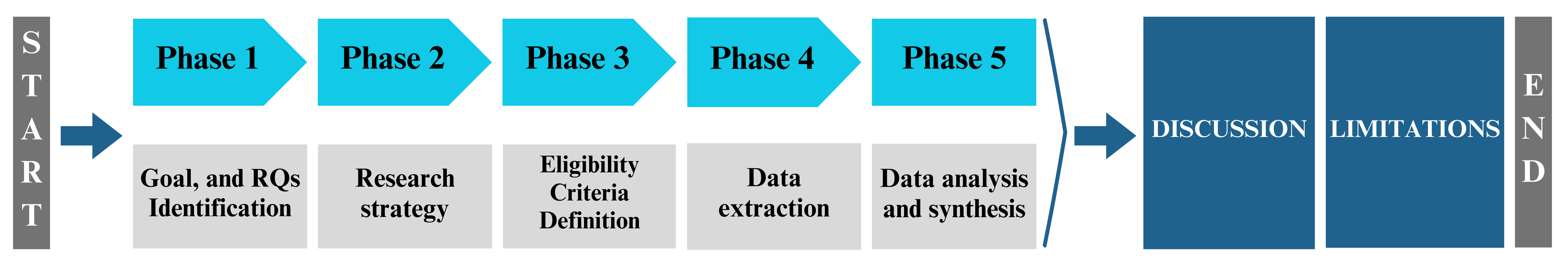}
    \caption{Research protocol used in the literature review}
    \label{fig:Researchprotocol}
\end{figure}

To perform this review, we followed the protocol proposed in \citep{10.1145/3210459.3210462}, and we completed the review process with the strategies presented in \citep{Kitchenham2007} for performing systematic literature reviews. The following subsections describe in detail the study design and its execution.
The literature review presented in this work was carried out through the following steps:
\begin{enumerate}
    \item \textbf{Goal and Research questions}: the goal and the correlated research questions were identified to guide the literature review;
    \item \textbf{Search strategy}: defining the strategy to collect previous works published in the literature, including research databases and query strings;
    \item \textbf{Eligibility criteria definition}: the criteria used to filter the collected studies have been defined;
    \item \textbf{Data extraction}: defining how relevant data were extracted to help answer the research questions;
    \item \textbf{Data analysis and synthesis}: defining how to organize extracted relevant data to answer the research questions.
\end{enumerate}
Fig. \ref{fig:Researchprotocol} summarizes the review protocol.

\subsection{Goal and Research Question Definition}
We formulated the following research questions to analyze the diverse dimensions of ChatGPT's impact.





Based on this goal, we defined the following research questions:
\begin{itemize}
    \item \textbf{RQ1}: What are the perceived  \textit{positive and negative impacts} of ChatGPT in contemporary society?
    \item \textbf{RQ2}: What are the \textit{emerging trends perceived} in ChatGPT development? 
    \item \textbf{RQ3}: Which \textit{areas of improvement} can be identified in the development of such technologies? 
\end{itemize}

\subsection{Search strategy}

The research has been split into two distinct parts: \textit{part one}, focused on grey literature, and \textit{part two}, focused on white literature.
Including both grey literature and academic contributions allowed us to conduct a deeper exploration of the impact of ChatGPT in various settings and from various perspectives.

In part one -- started on 29th November 2023 --  we focused on grey-literature sources, like blog posts and news articles from multiple domains -- such as business, education, technology, and society -- emphasizing ChatGPT. 
Here our goal was to feel the sentiment of the media and tech sphere and capture feelings on ChatGPT that cannot emerge from white-literature sources. Furthermore, some pilot searches on white-literature sources produced very few results, which did not allow us to derive reliable and consistent conclusions.

The string used for part one was:

\begin{tcolorbox}[colback=lightgray, colframe=lightgray, fontupper=\small]
(“ChatGPT” AND “social concerns”) 
(“ChatGPT” AND “social impact”)
(“ChatGPT” AND “Human rights”)
(“ChatGPT” AND “society*”)
(“ChatGPT” AND “education*”)
(“ChatGPT” AND “ethics”)
\end{tcolorbox}

This string was used to perform a keyword-based search on Google search engine\footnote{\url{https://www.google.it/}}; the search was performed in private-browsing mode, after logging out from personal accounts and erasing all web cookies and history \citep{Piasecki2017}.

In the end, this search resulted in 1230 literature sources.

While executing part one of the research, we continued executing periodic pilot searches for white-literature sources. 

In late February 2023, we noticed a notable change: the number of scientific articles addressing ChatGPT increased significantly, encompassing diverse approaches and perspectives. This may be due to the fact that academic papers need more time to be reviewed by peers and published (w.r.t. blog posts and news articles).

So, for part two of the research -- started on 22nd February 2023 -- we decided to use Google Scholar\footnote{\url{https://scholar.google.com/}} as white-literature search engine; here we searched for scientific articles of various fields -- such as business, education, technology, society, healthcare. 
The string used for part two of the research was the following:

\begin{tcolorbox}[colback=lightgray, colframe=lightgray, fontupper=\small]
("Large Language Model" AND "Social impact") ("Large Language Model" AND Human Rights") ("Large Language Model" OR "ChatGPT" AND "Ethics")("Large Language Model" AND "ChatGPT" AND "Ethics") ("Large Language Model" OR "ChatGPT" AND "Social concerns") ("Large Language Model'' AND "ChatGPT" AND "society*")  ("Large Language Model" AND "ChatGPT" AND "education*") ("Chat GPT" AND "social concerns") ("ChatGPT" AND "social impact") ("ChatGPT" AND "Human rights") ("ChatGPT" AND "society*") ("ChatGPT" AND "education*")  ("ChatGPT" AND "ethics") 
\end{tcolorbox}

In the end, this search resulted in 86 new literature sources.


All the documents obtained with this search strategy (both in part one and part two) were surveyed using a 3-stages information classification process.
In the first stage, only the title and keywords of the collected articles were read. In the second stage, we analyzed the abstract of each article while in the third stage we read the complete article. All these stages were conducted separately and in blind-view way by two of the authors. In case of a disagreement, a third author manually verified and took the final decision.

All found publications were subjected to the selection criteria outlined in Sec. \ref{subsec:selection_criteria} to determine their relevance for inclusion in the analysis.
 
\subsection{Eligibility Criteria Definition and Data Extraction}
\label{subsec:selection_criteria}

The selection procedure used for filtering the identified pool of 1316 papers was based on the following criteria: 
\begin{itemize}
  \item for every text the content should be mainly related to ChatGPT and LLM,
  \item the content should be written in English.
\end{itemize}
Then, the following criteria helped us to ensure we only included publications providing substantial information to our analysis, especially on ChatGPT, while we excluded papers with only brief mentions or tangential references:
\begin{itemize}
  \item For \textit{blog posts}, we required the author's name to be consistently provided, and the blogs should be specialized in the relevant subject matter.
  \item Regarding \textit{news articles}, we preferred those that offered an extensive analysis. We adopted this criterion because initial news coverage of ChatGPT tended to be repetitive, often focusing on its capabilities and limitations and just providing a brief history of the model.
  \item In the case of \textit{academic articles}, we sought diverse approaches to ensure a comprehensive perspective rather than lean solely on a single field, such as, for instance, the impact of ChatGPT in education.
\end{itemize}
After applying these selection criteria, we selected a total of \textbf{71} papers from our initial pool of \textbf{1316} articles.
 

In the Data Extraction step, we extracted all relevant data that could help answer any of the research questions.
The extraction process was performed by two of the authors and conflicts were solved by a third author in a blind-view way. 
We used Atlas.ti\footnote{\url{https://atlasti.com/}} to tabulate and organize data. More detailed information regarding the data and how it was indexed can be found in the online appendix \citep{online_appendix}.

\section{DATA ANALYSIS AND SYNTHESIS}
\label{sec:Dataanalysis}
Part one of the search has been conducted from 29th November 2022 to 22nd February 2023. Part two has been conducted from 22nd February 2023 to 19th May 2023. Table \ref{table:research_results} details the results obtained in both parts, as well as the documents selected once our selection criteria were applied.
\begin{table}[H]
\centering
\renewcommand{\arraystretch}{1.5}
\begin{tabular}{|>{\raggedright\arraybackslash}p{8 cm}|>{\raggedright\arraybackslash}p{2 cm}|>{\raggedright\arraybackslash}p{2 cm}|>{\raggedright\arraybackslash}p{2 cm}|}
\hline
\textbf{Research phase} & \textbf{Resources retrieved} & \textbf{Resources analyzed} & \textbf{Resources selected} \\
\hline
First part, until Feb. 22 on Google & 1230 & 300 & 25 \\
\hline
Second part, until May 19 on Google scholar & 86 & 63 & 46 \\
\hline
\end{tabular}
\caption{Amount of documents collected grouped by research phase.}
\label{table:research_results}
\end{table}

In order to answer RQ1, RQ2 and RQ3 we performed an analysis using Atlas.ti.
Atlas.ti is a qualitative research tool that enables the systematic organization of documentary resources by using codes and creating documentary categories. Atlas.ti allowed us to visualize and intuitively present content trends within the analyzed documents.
We employed this tool to analyze the selected papers, which were organized in Atlas.ti "document groups" following the categories in Table \ref{tab:coding}. 

\begin{table}[H]
\centering
\renewcommand{\arraystretch}{1.5}
\resizebox{\textwidth}{!}{
\begin{tabular}{|>{\raggedright\arraybackslash}p{3.5cm}|>{\raggedright\arraybackslash}p{3.5cm}|>{\raggedright\arraybackslash}p{3.5cm}|>{\raggedright\arraybackslash}p{3.5cm}|}
\hline
\multicolumn{4}{|c|}{\textbf{Codes for text analysis in Atlas.ti}} \\
\hline
\textbf{Positive Impact} & \textbf{Negative Impact} & \textbf{Emerging trends} & \textbf{Areas for improvement}\\
\hline
Benefits to education & Disinformation risk	 & Impact on the tech/AI market &  Need for appropriate regulation \\
\hline
Benefits to customer service & Negative impact on freedom of expression  & Copyright uncertainty & GDPR compliance concerns \\
\hline
Benefits of responses in real-time & Bias concerns &  Uncertainty over liability for production failures & Uncertainty of classification under the AI Act \\
\hline
... & ... & ... &\\
\hline
\end{tabular}
}
\caption{Codes for text analysis in Atlas.ti}
\label{tab:coding}
\end{table}

The categories and codes presented in Table 2 are derived from an in-depth analysis of the selected papers. The codes were primarily developed "in vivo," meaning they emerged as potential units of analysis during the reading and analysis process. For example, when repetitive references were made to the potential \textbf{positive impacts} of ChatGPT in education, we created the code "\textit{Benefits for education}". After reviewing and analyzing the documents, the codes were organized into four categories, which helped to address RQ1, RQ2, and RQ3. The criteria for categorization emerged from a thoughtful reflection on the codes and their contextual relevance.
In the category of positive impacts, codes such as "\textit{24/7 Availability}" and "\textit{Personalized feedback}" are included since they are frequently mentioned as strengths of ChatGPT. On the other hand, the category of \textbf{negative impacts} encompasses codes such as "\textit{Bias concerns}", which emerged as a recurring argument when discussing the potentially detrimental effects of this model. Other codes like "\textit{Privacy concern}" and "\textit{Water footprint}" were identified, further emphasizing the importance of addressing these issues in the context of ChatGPT's implementation.

In the category of \textbf{emerging trends}, we captured the unforeseen consequences, which, although potentially negative, arise unexpectedly from the evolution of the model itself. Examples include "\textit{Copyright uncertainty}" and the "\textit{Need to clarify private sector liability}", which pose challenges that trigger transformations in particular domains such as the "\textit{Impact on the tech/AI market}". This category also encompasses unexpected challenges to the AI Act \citep{aiact_proposal_2021} and its coverage of generative models. Another aspect within this category is "\textit{Unintentional misinformation}", referring to instances where the chat model provides unintentionally inaccurate information, a matter currently under scrutiny by various experts. Additionally, the category encompasses codes like "\textit{Skepticism about its actual impact}". These emerging trends shed light on the complex issues ChatGPT presents.

The final category encompasses \textbf{areas of improvement}, focusing on aspects that require further development to address all the adverse and unexpected effects of the model. For instance, the tag "\textit{Negative outcomes mitigation}" highlights the need for more efforts to minimize adverse consequences. The categorization of codes as "\textit{Uncertainty in data governance}" also responds to the criteria as an area of improvement rather than a negative impact. The above-mentioned decision was made considering that current legal frameworks do not adequately account for models like GPT, thus highlighting the need for specific measures to address these cases, which will likely be developed in the coming years. This category also encompasses improvement areas, such as "\textit{Limited up-to-date information}" and "\textit{Limited Medical terminology}", emphasizing the potential for enhancements.

\begin{figure} [H]
    \centering
    \includegraphics[scale=0.50]{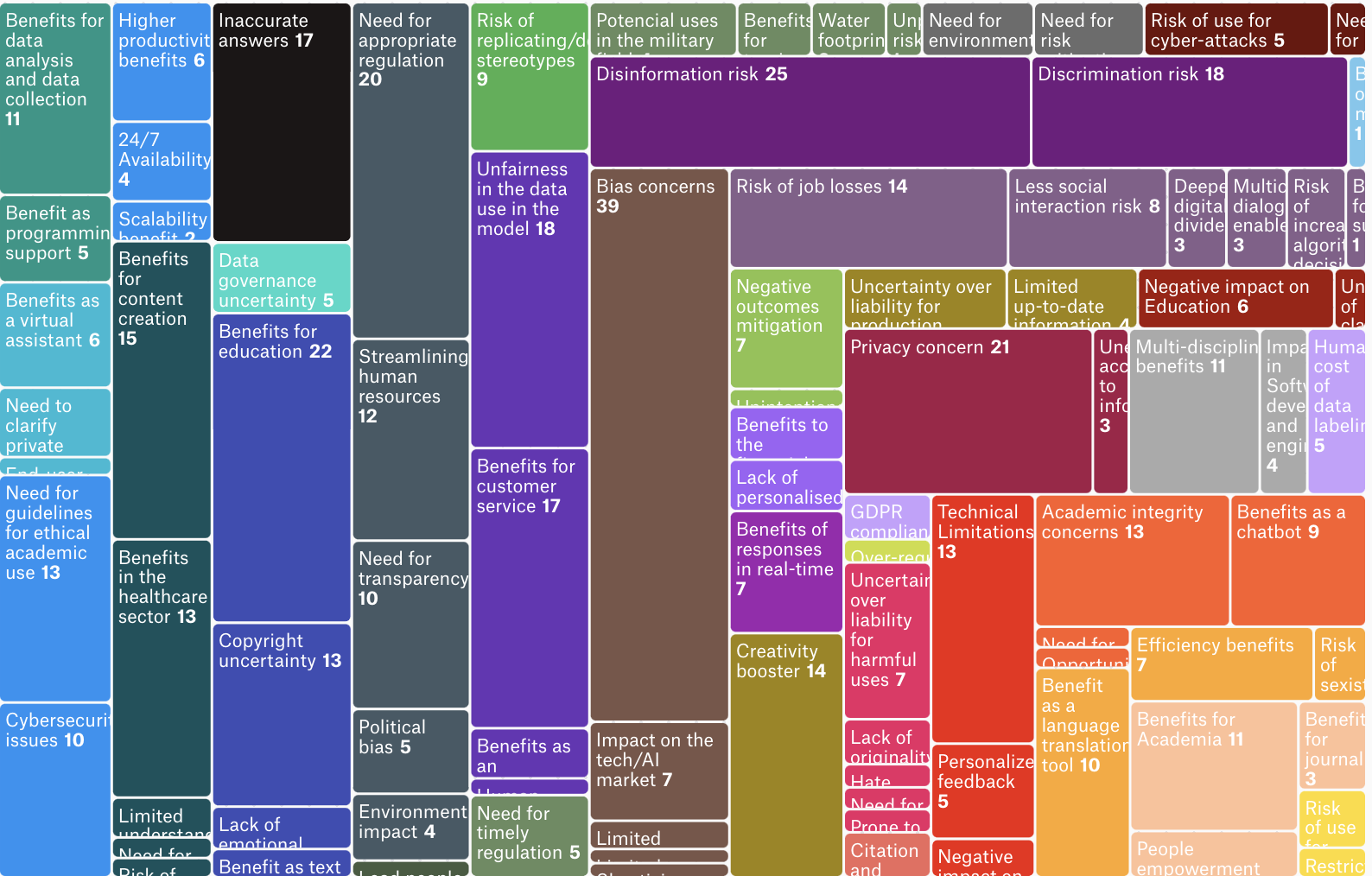}
    \caption{Treemap- Atlas.ti with codes distribution}
    \label{fig:Treemap}
\end{figure}

After establishing the codes and categories, we visually represented their distribution across the 71 documents. The resulting tree map in Atlas.ti, depicted in Figure \ref{fig:Treemap}, displays the frequency of the most repeated codes.

Table \ref{table:most_least_recurrent_codes} also shows the frequency of codes, highlighting the least and most recurrent codes within the Atlas.ti analysis.
\begin{table}[H]
\centering
\resizebox{\textwidth}{!}{
\begin{tabular}{|c|c|c|c|}
\hline
\textbf{Most recurrent} & \textbf{\#}     & \textbf{Least recurrent} & \textbf{\#}        \\ \hline
Bias concerns & 39                           & Unpredictability risk & 1                        \\ \hline
Disinformation risk & 25                     & Prone to injection attacks & 1                   \\ \hline
Benefits for education & 22                  & Over-regulation risk & 1                         \\ \hline
Privacy concern & 21                         & Opportunity to increase renewable energy use & 1 \\ \hline
Need for appropriate regulation & 20         & Need for using Renewable Energy Sources & 1      \\ \hline
Discrimination risk & 18                     & Need for human rights safeguards & 1             \\ \hline
Unfairness in the data use in the model & 18 & Need for explainability and traceability & 1     \\ \hline
Benefits for customer service & 17           & Need for Accessibility and Affordability & 1        \\ \hline
Inaccurate answers & 17                      & Limited Medical terminology & 1                  \\ \hline
Benefits for content creation & 15           & Lead people into extremist positions risk & 1    \\ \hline
\end{tabular}
}
\caption{Most and least recurrent codes; each category is associated with its number of occurrences.}
\label{table:most_least_recurrent_codes}
\end{table}

\begin{figure}[H]
    \centering
    \includegraphics[scale=0.38]{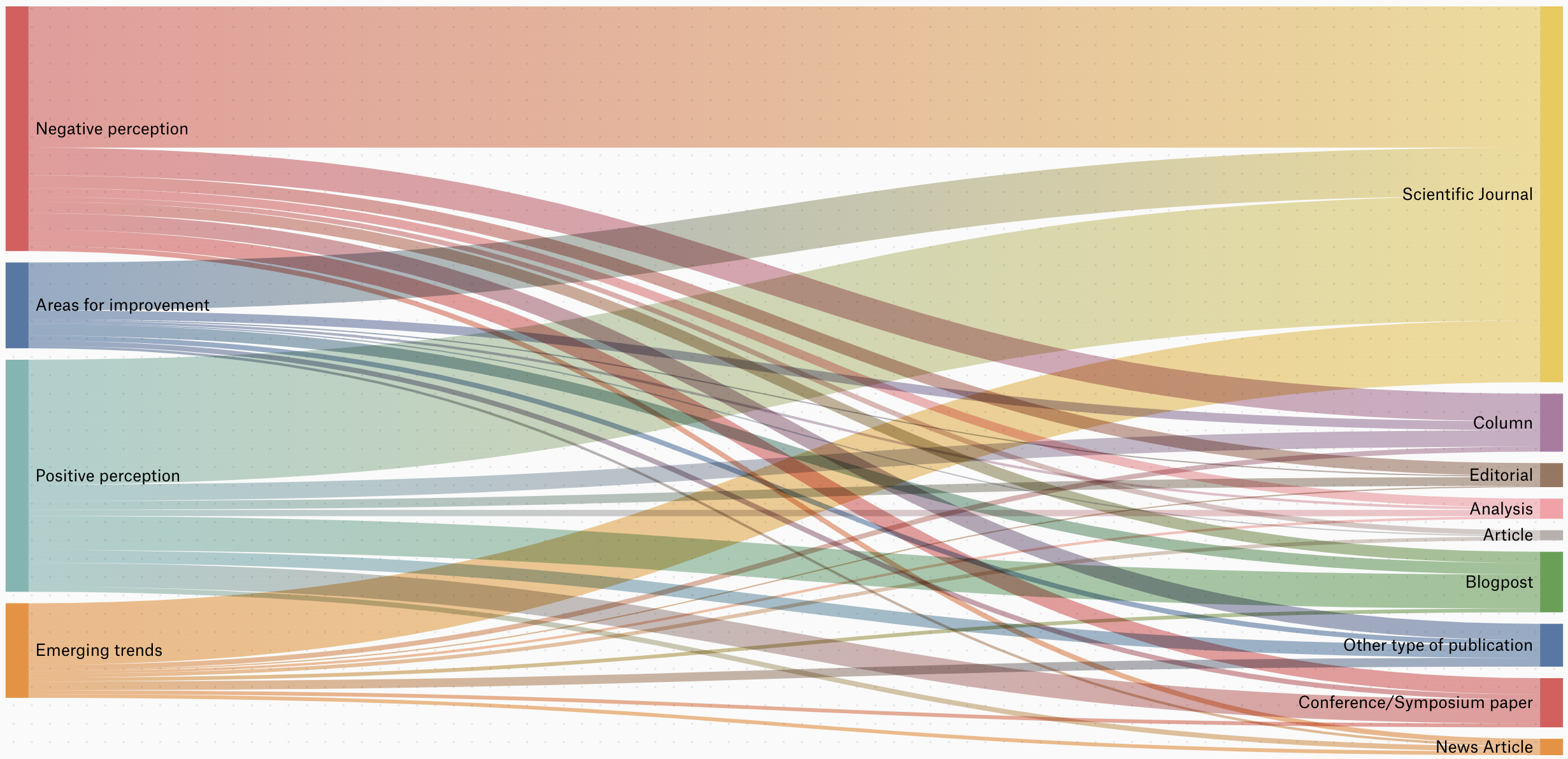}
    \caption{Sankey diagram illustrating the distribution of code groups across document groups}
    \label{fig:Sankey}
\end{figure}

In addition, Fig.\ref{fig:Sankey} displays a Sankey diagram that graphically depicts the distribution of code categories across the analyzed documents (in document groups-unit). 
The diagram shows that scientific papers, blog posts, conference symposiums, and other types of publications encompass all coding groups (negative and positive impacts, areas of improvement, and emerging trends). See Table A2 in online appendix \citep{online_appendix} for a description of document categories. The articles category includes just emerging trends, areas for improvement, and negative perceptions. Most document categories, including scientific papers, columns, analyses, editorials, and articles, exhibit a negative tendency. Notably, positive perceptions are more prevalent in blog-posts, conference and symposium papers, and to a lesser degree in news articles. "\textit{Bias concern}" and "\textit{Disinformation risk}" are two of the most common codes contributing to negative perceptions. In contrast, "\textit{Benefits for customer service}" and "\textit{Multidisciplinary benefits}" are the most prevalent codes in the documents with a positive trend. Table A3 in online appendix \citep{online_appendix} details tendencies and code frequency across all document categories.

\section{Discussion}

\textbf{RQ1.} \textit{What are the perceived  positive and negative impacts of ChatGPT in contemporary society?}

Throughout our literature review, we identified several positive and negative impacts attributed to ChatGPT. Noteworthy benefits include:
\textbf{the potential for enhancing customer service;} multiple papers emphasize the positive impact of ChatGPT in this domain \citep{international_journal_of_human_rights_law_review_chatgpt_2023, ray_chatgpt_2023,paul_chatgpt_2023,hillemann_chatgpt_2023, davis_chatgpt_2022, lock_what_2022,deo_is_2023, rivas_marketing_2023, kumordzie_all_2023, levente_pros_2023, marr_what_2023, abdullah_chatgpt_2022, khowaja_chatgpt_2023, gupta_chatgpt_2023, bruff_teaching_2023,iskender_holy_2023}. The model is highlighted as an enabler of cross-cultural dialogue, facilitating communication between individuals from different cultural backgrounds \citep{international_journal_of_human_rights_law_review_chatgpt_2023}. Moreover, ChatGPT offers the advantage of \textbf{automating repetitive tasks}, freeing time for more complex and value-added activities \citep{davis_chatgpt_2022, levente_pros_2023, khan_chatgpt_2023,gupta_chatgpt_2023}. These benefits extend to various sectors, including business and healthcare \citep{deo_is_2023}.
Another key advantage is its \textbf{availability around the clock}. This 24/7 accessibility proves valuable in commercial, healthcare, and educational contexts \citep{paul_chatgpt_2023, levente_pros_2023, lee_rise_2023, sun_chatgpt_2023, cardoso_we_2023}. The model's continuous availability ensures timely assistance and support, covering users' diverse needs.

ChatGPT also demonstrates significant \textbf{potential in education}, offering various advantages \citep{marr_what_2023,shidiq_use_2023,lee_chatgpt_2023, sun_chatgpt_2023,mhlanga_open_2023,rivas_marketing_2023, abdullah_chatgpt_2022, bahrini_chatgpt_2023,ray_chatgpt_2023,ausat_can_2023, solaiman_release_2019, lee_chatgpt_2023, tiunova_chatgpt_2023, iskender_holy_2023, bruff_teaching_2023,geertsema_chatgpt_2023}. Despite concerns about plagiarism and academic integrity, the model's integration can enhance teaching practices in several ways. It can, for example, automates curriculum creation, enabling educators to save time and streamline the process \citep{marr_what_2023, bahrini_chatgpt_2023}. Moreover, it facilitates the development of innovative educational content, fostering an engaging learning environment \citep{lee_rise_2023, shidiq_use_2023}. Additionally, the model serves as a personalized study support assistant, providing tailored guidance and assistance to individual learners \citep{ray_chatgpt_2023, rivas_marketing_2023,sun_chatgpt_2023, mhlanga_open_2023, abdullah_chatgpt_2022, ausat_can_2023}.
In this perspective, a vision emphasizes the need for controlled integration and adherence to academic guidelines to ensure responsible and ethical use of generative AI models in education \citep{sun_chatgpt_2023}. By establishing appropriate regulations and ethical frameworks, the educational benefits of ChatGPT can be maximized while addressing concerns related to plagiarism and promoting an enriching learning experience.

In the medical field, the model has shown promise in research, data analysis, and telemedicine applications, \textbf{contributing to advancements in healthcare} \citep{bahrini_chatgpt_2023}. Furthermore, diverse papers recognize its potential contribution to \textbf{addressing environmental challenges}. For instance, it may help find innovative solutions to reduce water consumption in the AI industry, highlighting its role in promoting sustainability \citep{george_environmental_2023}.
Similarly, ChatGPT is also appreciated for its potential as an \textbf{informative and accountability tool within institutions} \citep{ray_chatgpt_2023, cardoso_we_2023, biswas_role_2023}.  Lastly, it positively impacts journalism, where it can help with content creation, fact-checking, and generating engaging narratives \citep{davis_chatgpt_2022, marr_what_2023, bahrini_chatgpt_2023}. 

Conversely, the most recurrent social concern is \textbf{bias}. Several papers \citep{international_journal_of_human_rights_law_review_chatgpt_2023, ray_chatgpt_2023, paul_chatgpt_2023, lee_rise_2023, shidiq_use_2023, biddle_internets_2023, c_chatgpt_2023, equality_now_chatgpt-4_2023, robson_ai_2023, perrigo_exclusive_2023,li_dark_2023, biswas_role_2023,vidhya_prognosis_2023, khan_chatgpt_2023, ausat_can_2023, paul_chatgpt_2023,solaiman_release_2019,tamkin_understanding_2021,treude_she_2023, farrell_spirals_2022, tiunova_chatgpt_2023, thirunavukarasu_trialling_2023, stepanechko_english_2023, geertsema_chatgpt_2023} discuss the risk of deepening existing biases and how ChatGPT can include sexist and racist views due to the characteristics of the data used during its training. In this sense, there is a special concern about using these tools in various sectors, including finance \citep{khan_chatgpt_2023} and other social activities, which can replicate and deepen structural and historical inequalities. 
Our analysis also reveals another perceived negative impact, which is its potential to generate \textbf{false information} and facilitate the spread of disinformation \citep{ray_chatgpt_2023, paul_chatgpt_2023, hillemann_chatgpt_2023, davis_chatgpt_2022,lock_what_2022, bahrini_chatgpt_2023,equality_now_chatgpt-4_2023,robson_ai_2023,li_dark_2023,biswas_role_2023, vidhya_prognosis_2023, khan_chatgpt_2023,helberger_chatgpt_2023, air_ethics_2023, wolf_ai_2023, vallance_chatgpt_2022, rozado_political_2023, paul_chatgpt_2023, solaiman_release_2019,tamkin_understanding_2021, farrell_spirals_2022, tiunova_chatgpt_2023, khowaja_chatgpt_2023}. This misuse of technology directly affects rights related to access to accurate information and freedom of expression, as well as democratic stability.
This phenomenon is particularly relevant as, although remarkable in artificial intelligence, the advancements and improvements in generative models have raised concerns and uncertainties regarding their \textbf{impact on democratic processes}. There is a special concern about the ease with which false information can be generated and disseminated, especially in critical contexts such as elections, referendums, political instability, war conflicts, or under dictatorial regimes.
Furthermore, this potential negative impact includes the digital public sphere, where spreading \textbf{hate speech} on social networks \citep{institute_for_human_rights_and_business_we_2023} can lead to social fragmentation and bolster the manipulation of democratic institutions and social control. False information can be weaponized across various domains, from the stock market to information warfare and propaganda.

In the same vein, another relevant concern is \textbf{privacy} \citep{international_journal_of_human_rights_law_review_chatgpt_2023,ray_chatgpt_2023,paul_chatgpt_2023,deo_is_2023,lee_rise_2023,bahrini_chatgpt_2023,mhlanga_open_2023,c_chatgpt_2023,biswas_role_2023,robson_ai_2023,perrigo_exclusive_2023, li_dark_2023,biswas_prospective_2023,vidhya_prognosis_2023,khan_chatgpt_2023, helberger_chatgpt_2023,vallance_chatgpt_2022,khlaif_ethical_2023, paul_chatgpt_2023, tiunova_chatgpt_2023, khowaja_chatgpt_2023, iskender_holy_2023}. 
Different aspects contribute to this concern, including the coverage of these models under AI Act regulations, its extensive use of user data (particularly minors), the potential for surveillance applications, the data vulnerability to cyber attacks targeting ChatGPT, and the privacy implications when integrating the model into various domains such as education and the military.
Privacy is a growing concern as illustrated by the case of the Italian state ban. In this case, the Italian authorities asked OpenAI to “expand its privacy policy for users and made it also accessible from the sign-up page prior to registration with the service” \citep{garante_per_la_protezione_dei_dati_personali_chatgpt_2023} in order to operate in Italy. This case highlights the urgency for ensuring responsible and transparent use of the technology.
Another important negative impact is \textbf{the risk of job loss}, as automation and AI capabilities advance \citep{davis_chatgpt_2022,deo_is_2023,rivas_marketing_2023,lee_chatgpt_2023,biddle_internets_2023,robson_ai_2023, khan_chatgpt_2023, air_ethics_2023,wolf_ai_2023, curtis_chatgpt_2023, gabbiadini_does_2023, iskender_holy_2023, stepanechko_english_2023}.

Furthermore, other papers show apprehension about \textbf{over-regulation} \citep{helberger_chatgpt_2023}, highlighting the needed balance between ensuring ethical and responsible use of AI technologies while avoiding stifling innovation. Additionally, the model's own cybersecurity is listed as a negative impact \citep{levente_pros_2023, bahrini_chatgpt_2023,c_chatgpt_2023,khan_chatgpt_2023,helberger_chatgpt_2023,air_ethics_2023,khowaja_chatgpt_2023}. The comprehensive list of negative perceptions on ChatGPT impacts can be found in Table A1 in online appendix \citep{online_appendix}, providing further insights into the concerns found in the literature. 

\textbf{RQ2.} \textit{What are the emerging trends perceived in ChatGPT development?}

An essential part of our review shows emerging trends \citep{hillemann_chatgpt_2023,tech_telegraph_microsoft_2023,dibenedetto_chatgpts_2023, deo_is_2023,equality_now_chatgpt-4_2023,institute_for_human_rights_and_business_we_2023, kumordzie_all_2023, vallance_chatgpt_2022, perrigo_exclusive_2023, al_ashry_chat_2023, bareis_we_2023, curtis_chatgpt_2023,rutinowski_self-perception_2023, lee_chatgpt_2023, rozado_political_2023, george_environmental_2023, levente_pros_2023, sun_chatgpt_2023, biswas_prospective_2023, vidhya_prognosis_2023, suguri_motoki_more_2023, rivas_marketing_2023, khlaif_ethical_2023, bahrini_chatgpt_2023,ray_chatgpt_2023, khan_chatgpt_2023, ausat_can_2023, paul_chatgpt_2023, tamkin_understanding_2021, solaiman_release_2019, lee_chatgpt_2023, geertsema_chatgpt_2023, tiunova_chatgpt_2023, khowaja_chatgpt_2023, iskender_holy_2023, stepanechko_english_2023, bruff_teaching_2023}.
Within this category we found \textbf{uncertainty surrounding copyright}, which is a prominent trend, raising doubts about the possibility of profiting from chat-generated content and whether such content is subject to copyright protection. Resolving these issues requires legislative intervention, leading to a particular and complex debate among authorities regarding the legal implications of AI-generated content \citep{hillemann_chatgpt_2023}. There is also a growing demand for the \textbf{enhancement of regulatory frameworks} to safeguard original content, considering that models like ChatGPT have the potential to negatively impact the work of scientists, writers, researchers, and artists \citep{al_ashry_chat_2023, ray_chatgpt_2023, tiunova_chatgpt_2023, khowaja_chatgpt_2023, iskender_holy_2023, stepanechko_english_2023}. Notably, analyses such as by Khowaja et al. \citep{khowaja_chatgpt_2023} raise crucial questions regarding ownership rights over the data used to train the model and the ownership of the model itself.
As mentioned earlier, there is a dominant call for establishing \textbf{ethical use guidelines}, both at a general societal level \citep{gabbiadini_does_2023} and specifically within educational and research institutions \citep{mhlanga_open_2023, khlaif_ethical_2023, ausat_can_2023, solaiman_release_2019, tamkin_understanding_2021, lee_chatgpt_2023, tiunova_chatgpt_2023, gabbiadini_does_2023, iskender_holy_2023, stepanechko_english_2023, bruff_teaching_2023}. These guidelines will be pivotal in ensuring responsible adoption models such as ChatGPT.

Another emerging trend is developing \textbf{transparency mechanisms} \citep{equality_now_chatgpt-4_2023, lee_chatgpt_2023, lee_rise_2023,bahrini_chatgpt_2023, ray_chatgpt_2023, khan_chatgpt_2023, tiunova_chatgpt_2023, khowaja_chatgpt_2023,stepanechko_english_2023}. Transparency is considered a vital strategy to address resistance toward adopting these models in various contexts \citep{bahrini_chatgpt_2023} and to mitigate potential AI stigmatization. However, it is also acknowledged that transparency poses challenges that need to be overcome \citep{khowaja_chatgpt_2023}.

Another crucial issue to highlight is the \textbf{accountability for potentially harmful uses} of such technology \citep{deo_is_2023, biswas_prospective_2023, vidhya_prognosis_2023, ray_chatgpt_2023, paul_chatgpt_2023, khowaja_chatgpt_2023, stepanechko_english_2023, helberger_chatgpt_2023}, which involves both the end-users and the companies responsible for its development \citep{helberger_chatgpt_2023, deo_is_2023, institute_for_human_rights_and_business_we_2023, george_environmental_2023, biswas_prospective_2023, vidhya_prognosis_2023, ray_chatgpt_2023,paul_chatgpt_2023, khowaja_chatgpt_2023, stepanechko_english_2023}. This concern extends to applications in sensitive domains like the military \citep{deo_is_2023}.
Furthermore, other several notable trends emerge, including the need for \textbf{timely and appropriate regulation} \citep{gabbiadini_does_2023, bruff_teaching_2023}, the existence of \textbf{political bias} (29, 40, 42, 62), \textbf{transformations within the AI market} \citep{deo_is_2023, equality_now_chatgpt-4_2023, kumordzie_all_2023,perrigo_exclusive_2023, tamkin_understanding_2021, farrell_spirals_2022, geertsema_chatgpt_2023}, and the \textbf{integration of renewable technologies and environmental awareness} within this field \citep{c_chatgpt_2023, international_journal_of_human_rights_law_review_chatgpt_2023}.

\textbf{RQ3.} \textit{Which areas of improvement can be identified in the development of such technologies?} 

The findings concerning areas for improvement within the field of generative AI present a diverse range of perspectives \citep{hillemann_chatgpt_2023,helberger_chatgpt_2023, deo_is_2023, equality_now_chatgpt-4_2023, biddle_internets_2023, institute_for_human_rights_and_business_we_2023, weetech_solution_what_2023, levente_pros_2023, air_ethics_2023, kumordzie_all_2023, wolf_ai_2023, vallance_chatgpt_2022, perrigo_exclusive_2023, al_ashry_chat_2023, levente_pros_2023, sun_chatgpt_2023, biswas_role_2023, zhuo_exploring_2023, abdullah_chatgpt_2022, bahrini_chatgpt_2023, ray_chatgpt_2023, khan_chatgpt_2023, paul_chatgpt_2023,sobieszek_playing_2022, solaiman_release_2019, tamkin_understanding_2021, c_chatgpt_2023, tiunova_chatgpt_2023, khowaja_chatgpt_2023,gabbiadini_does_2023, gupta_chatgpt_2023, iskender_holy_2023, stepanechko_english_2023, bruff_teaching_2023}. One prominent area is the examination of regulations \citep{hillemann_chatgpt_2023, helberger_chatgpt_2023, deo_is_2023, equality_now_chatgpt-4_2023, biddle_internets_2023, institute_for_human_rights_and_business_we_2023, kumordzie_all_2023, wolf_ai_2023, al_ashry_chat_2023, george_environmental_2023, levente_pros_2023, lee_chatgpt_2023, ray_chatgpt_2023, solaiman_release_2019, tamkin_understanding_2021, tiunova_chatgpt_2023, gabbiadini_does_2023, bruff_teaching_2023}, particularly regarding whether the risk-based approach outlined in the \textbf{AI Act} effectively covers generative models \citep{helberger_chatgpt_2023, wolf_ai_2023}. It is suggested that comprehensive guidelines should encompass the entire spectrum, from its application to the AI Research and Development (R\&D) \citep{george_environmental_2023}. Furthermore, there is a growing advocacy for \textbf{a people-centred vision} \citep{wolf_ai_2023, kumordzie_all_2023, ray_chatgpt_2023} that emphasizes the importance of human rights and ethical considerations in designing and implementing generative AI systems.
Specifically, Solaiman et al. \citep{solaiman_release_2019} examine the need to "build frameworks for navigating trade-offs" and develop decision-making frameworks that account for the complexities and potential trade-offs associated with generative AI. Similarly, Li \citep{levente_pros_2023} highlights the necessity to transform the European regulatory paradigm to effectively address the challenges posed by LLMs.

Another area of opportunity lies in addressing \textbf{technical limitations} within generative AI models \citep{levente_pros_2023, curtis_chatgpt_2023, sun_chatgpt_2023, abdullah_chatgpt_2022, bahrini_chatgpt_2023, cao_comprehensive_2023, paul_chatgpt_2023, sobieszek_playing_2022, tamkin_understanding_2021, tiunova_chatgpt_2023, thirunavukarasu_trialling_2023,gupta_chatgpt_2023, iskender_holy_2023}. For instance, a significant challenge is the presence of \textbf{fictional references} \citep{tiunova_chatgpt_2023} within the generated text, which hampers its reliability. Additionally, \textbf{words repetition} further affects the produced text's overall quality \citep{wolf_ai_2023}. Moreover, the phenomenon of inaccurate information, named "\textit{hallucinations}" \citep{wolf_ai_2023, tamkin_understanding_2021}, and the lack of context understanding \citep{paul_chatgpt_2023}, are also identified as areas requiring attention and improvement.
Furthermore, the literature highlights the need for enhanced \textbf{risk mitigation mechanisms} \citep{equality_now_chatgpt-4_2023, biddle_internets_2023, weetech_solution_what_2023, levente_pros_2023, air_ethics_2023, vallance_chatgpt_2022, perrigo_exclusive_2023}. This entails \textbf{refining processes for filtering potentially harmful responses} \citep{biddle_internets_2023}, \textbf{improving the quality and reliability of the data used to train the models} \citep{levente_pros_2023}, and \textbf{incorporating ethical guidelines} for its development \citep{air_ethics_2023}, among other measures \citep{perrigo_exclusive_2023}.

\textbf{Data governance} is another significant area that requires attention \citep{deo_is_2023, ray_chatgpt_2023, khan_chatgpt_2023, tiunova_chatgpt_2023}; this crucial task involves safeguarding sensitive information against security breaches, unauthorized access, and information theft \citep{ray_chatgpt_2023, khan_chatgpt_2023}. In the same vein, \textbf{establishing clear guidelines regarding the scope and limitations of information exchange with third parties} is also paramount \citep{tiunova_chatgpt_2023}.

Other areas of opportunity include \textbf{the need for up-to-date data} \citep{sun_chatgpt_2023,zhuo_exploring_2023, abdullah_chatgpt_2022, gupta_chatgpt_2023} to ensure the accuracy and relevance of generative AI models. However, this requirement presents a trade-off between incorporating new data to improve performance or addressing data governance issues first.

The literature review also highlights several other areas of opportunity for improvement; these include promoting \textbf{end-user responsibilities} \citep{li_dark_2023}, advocating for \textbf{timely regulation} \citep{li_dark_2023, tiunova_chatgpt_2023}, and \textbf{raising awareness of environmental impact} \citep{tiunova_chatgpt_2023, geertsema_chatgpt_2023}, among other considerations.

\section{CONCLUSION AND FUTURE WORK}
\label{sec:conclusion}
While we are still in the early stages of evaluating the social impact of generative AI models, this systematic literature review allows us to gain initial insights into the perceptions surrounding their emergence in contemporary society. Our analysis has revealed notable areas of concern, particularly \textit{privacy} and the \textit{potential for bias}. As generative models continue to be adopted in diverse social contexts, addressing and mitigating issues related to inequality, bias, discrimination, and stereotypes becomes urgent.

In light of this, it is essential to note that generative AI reflects the social context in which it was created. As these models are trained on data that captures various aspects of our reality, it becomes clear that addressing their flaws and biases requires a comprehensive understanding of the broader social context within which they operate. Rather than solely focusing on repairing the model, it is imperative to also engage in a critical examination of the social factors that contribute to these biases and limitations.

Some analyses argue that generative AI models, such as ChatGPT, are not intended to address social inequalities. While this may be true, it is also essential that these models do not inadvertently contribute to exacerbating social issues. Acknowledging that scientific breakthroughs do not occur within a social vacuum is critical. Therefore, we must foster a conscious, responsible, and ethically-driven progression of generative AI. 
Equally important is to emphasize that generative AI models hold immense potential and offer substantial benefits across various fields and sectors, including education, medicine, marketing, business, research, and science. Their impact extends beyond innovation and significantly influences the legislative landscape. Consequently, policymakers need to address the necessity for appropriate regulation that not only addresses significant concerns associated with their use, 
but also supports and facilitates the ethical development of generative AI models \citep{clayton_sam_2023, anthropic_anthropic_nodate}.
Undoubtedly, AI generative models have reshaped our way of being in the world, triggering profound changes in our perception and engagement with it. In our relentless pursuit to emulate human interaction, we have also confronted stereotypes, biases, and imperfections. Rather than succumbing to discouragement, we should use them as motivation to address them diligently and strive for continuous improvement. More work is required to develop more robust frameworks and ethical guidelines, not only to improve accuracy and efficiency but also to ensure responsible deployment.
As part of future work, we propose evaluating ChatGPT regulation in the US, Europe, and Latin America. This analysis will examine how current legal tools address generative models' challenges in particular locations.
Additionally, to understand ChatGPT users' professional and social views, a survey has been designed and will be distributed among professionals and researchers from diverse universities and research centers worldwide. 
With these exercises, we want to gain a comprehensive understanding of its adoption, regulatory challenges, user perspectives, and deepening into its social impact.\\










\bibliographystyle{unsrtnat}
\bibliography{references}  






\end{document}